\documentclass[10pt,twocolumn,letterpaper]{article}

\usepackage{cvpr}
\usepackage{times}
\usepackage{epsfig}
\usepackage{graphicx}
\usepackage{amsmath}
\usepackage{amssymb}

\usepackage[noend]{algpseudocode}
\usepackage{algorithmicx,algorithm}
\usepackage{multirow}
\usepackage{subfigure}

\usepackage[pagebackref=true,breaklinks=true,letterpaper=true,colorlinks,bookmarks=false]{hyperref}

\cvprfinalcopy 

\def\cvprPaperID{2721} 

\ifcvprfinal\fi
\begin{document}

\title{Scale-Aware Face Detection}

\author{Zekun Hao$^{1}$, Yu Liu$^{1}$, Hongwei Qin$^{2}$, Junjie Yan$^2$, Xiu Li$^2$, Xiaolin Hu$^2$ \\
$^1$SenseTime, $^2$Tsinghua University\\
\tt\small \{haozekun, yanjunjie\}@outlook.com, liuyuisanai@gmail.com, \\
\tt\small \{qhw12@mails., xlhu@, li.xiu@sz.\}tsinghua.edu.cn
}

\maketitle

\begin{abstract}

Convolutional neural network (CNN) based face detectors are inefficient in handling faces of diverse scales. 
They rely on either fitting a large single model to faces across a large scale range or multi-scale testing. Both are computationally expensive.
We propose Scale-aware Face Detection (SAFD) to handle scale explicitly using CNN, and achieve better performance with less computation cost. 
Prior to detection, an efficient CNN predicts the scale distribution histogram of the faces. Then the scale histogram guides the zoom-in and zoom-out of the image. Since the faces will be approximately in uniform scale after zoom, they can be detected accurately even with much smaller CNN. 
Actually, more than 99\% of the faces in AFW can be covered with less than two zooms per image. Extensive experiments on FDDB, MALF and AFW show advantages of SAFD.

\end{abstract}

\section{Introduction}

Face detection is one of the most widely used computer vision applications. Popular face detectors have been proposed, including the Viola-Jones\cite{viola2004robust}and its extensions, part model \cite{felzenszwalb2010object} and its successors and the convolutional neural network (CNN) based approaches \cite{vaillant1994original}. The CNN based approaches have recently shown great successes~\cite{huang2015densebox,yang2015facial,chen2016supervised}.

A face detection system should be able to handle faces of various scales, poses and appearances.
For CNN-based face detectors, the variance in pose and appearance can be handled by the large capacity of convolutional neural network. The variance in scale, however, is not carefully considered and there is room for improvement. The popularity of CNN in computer vision domain largely comes from its translation invariance property, which significantly reduces computation and model size compared to fully-connected neural networks. However, as for scale invariance, CNN meets the limitation
that is similar to the limitation of translation invariance for fully-connected networks. The CNN does not inherently have scale invariance. A CNN can be trained to have certain extent of scale invariance, but it needs more parameters and more complex structures to retain performance. Despite the importance, works that involve scale are rarely seen, and no work focuses on the essence of the scale problem. One possible reason is that in academic research, the simple multi-scale testing on
image pyramids can be used to avoid the problem and get good accuracy. However, multi-scale testing leads to heavy computation cost. Another way to avoid this problem is to fit a CNN model to multiple scales. This may also lead to an increase in model size and computation. 

\begin{figure}[t]
\begin{center}
  \includegraphics[width=\linewidth]{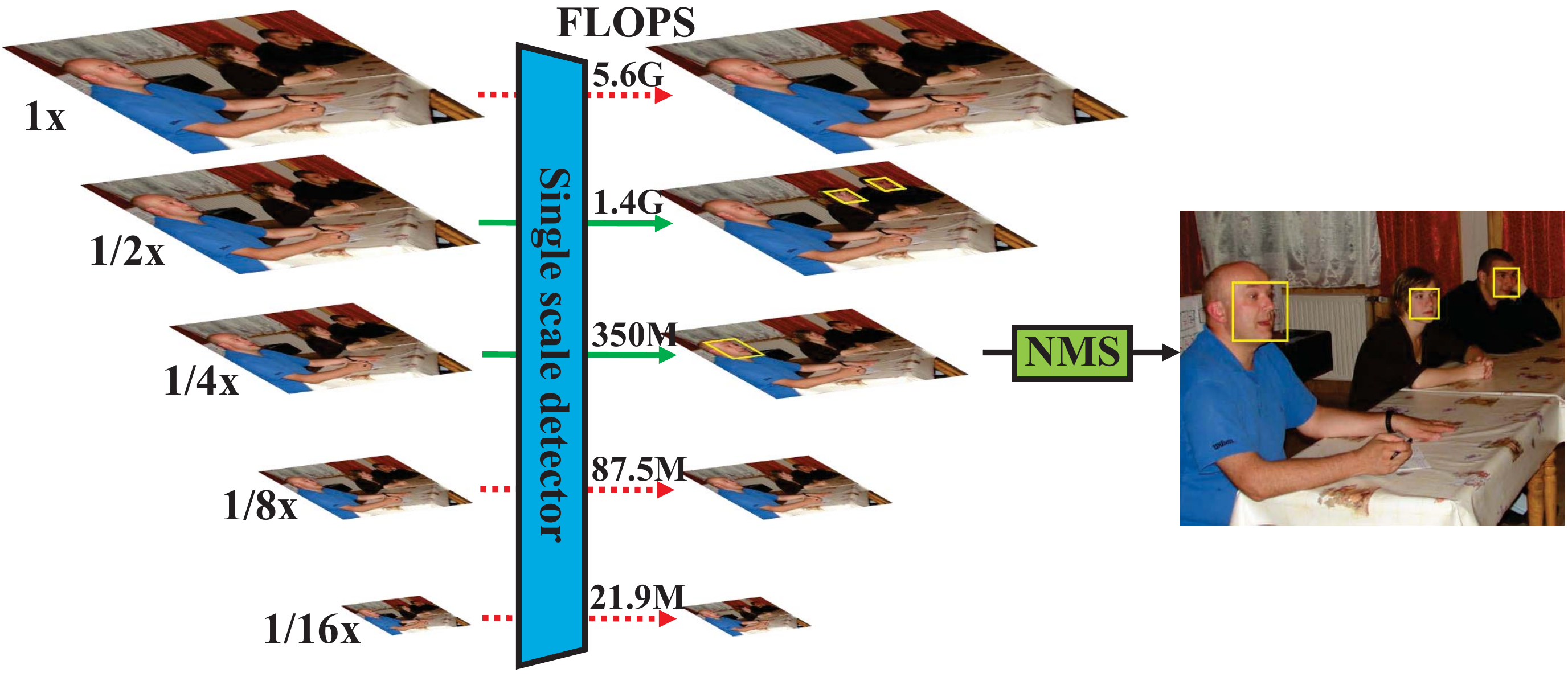}
\end{center}
   \caption{\textbf{The motivation of SAFD. } Single-scale detectors need to perform multi-scale testing on image pyramids in order to cover a large scale range. However, in most cases only a few layers in the image pyramids contain faces of valid scales (green arrow). Finding faces on those invalid scales is a waste of computation (red dashed arrow). In the proposed method, we show that the prediction of those valid scales can be done efficiently by a CNN, which considerably reduces computation.}
\label{fig:motivation}
\end{figure}

To solve this problem, we consider estimating the scale explicitly. If we know the face scales in each image, we can resize the image to suitable scales that best fit the detector. It eliminates the need to cover variances caused by scales so that smaller detector network can be used while achieving even better performance. It also prevents exhaustively testing all the scales in an image pyramid, which saves computation, as illustrated in Figure~ \ref{fig:motivation}.

In this way, the face detection procedure can be divided into face scale estimation and single scale detection. 

The scale proposal stage is implemented through a light-weight, fully-convolutional network called \textit{Scale Proposal Network (SPN)}. The network can generate a global face scale histogram from an input image of arbitrary sizes. A global max-pooling layer is placed at the end of the network, so it outputs a fixed-length vector regardless to the size of input image. The histogram vector encodes the probability of the existence of faces at certain scales.  
The input image is resized according to the histogram to ensure all the faces are within the valid range of the following detection stage. The SPN can be trained with the image-level supervision of ground truth histogram vectors and no face location information is required. 

The second stage is single-scale face detection. The face scales  of the training images have already been normalized to a narrow range prior to detection, so a simple detector that covers a narrow scale range can achieve high performance. We use a Region Proposal Network(RPN) as the detector in all the experiments because it is simple, fast and accurate on face detection task because there is only one object class.

By using the two-stage SA-RPN method, the average computation cost can be reduced while achieving state-of-the-art accuracy. The reasons are two-fold. On one hand, the single-scale detector adopts a smaller network than a multi-scale detector. Experiments show that a small network performs better if it only focuses on faces within a narrow scale range.
On the other hand, when a face occupies a large part of the image, it can be down-sampled to save computation in detection. When a face is smaller than the optimal range, up-sampling makes it easier to be detected.

\vspace{10pt}
\noindent \textbf{Contributions. } The contributions are in the following:
\begin{enumerate}
\setlength{\itemsep}{0ex}
\item We propose to divide face detection problem into two sub-problems: scale estimation and single-scale detection. Both problems are cheap in computation and overall computation is reduced while achieving state-of-the-art performance on FDDB, MALF and AFW.
\item We introduce SPN for generating fine-grained scale proposals and the network can be trained easily via image-level supervision.
\end{enumerate}

\begin{figure*}
\begin{center}
  \includegraphics[width=0.8\linewidth]{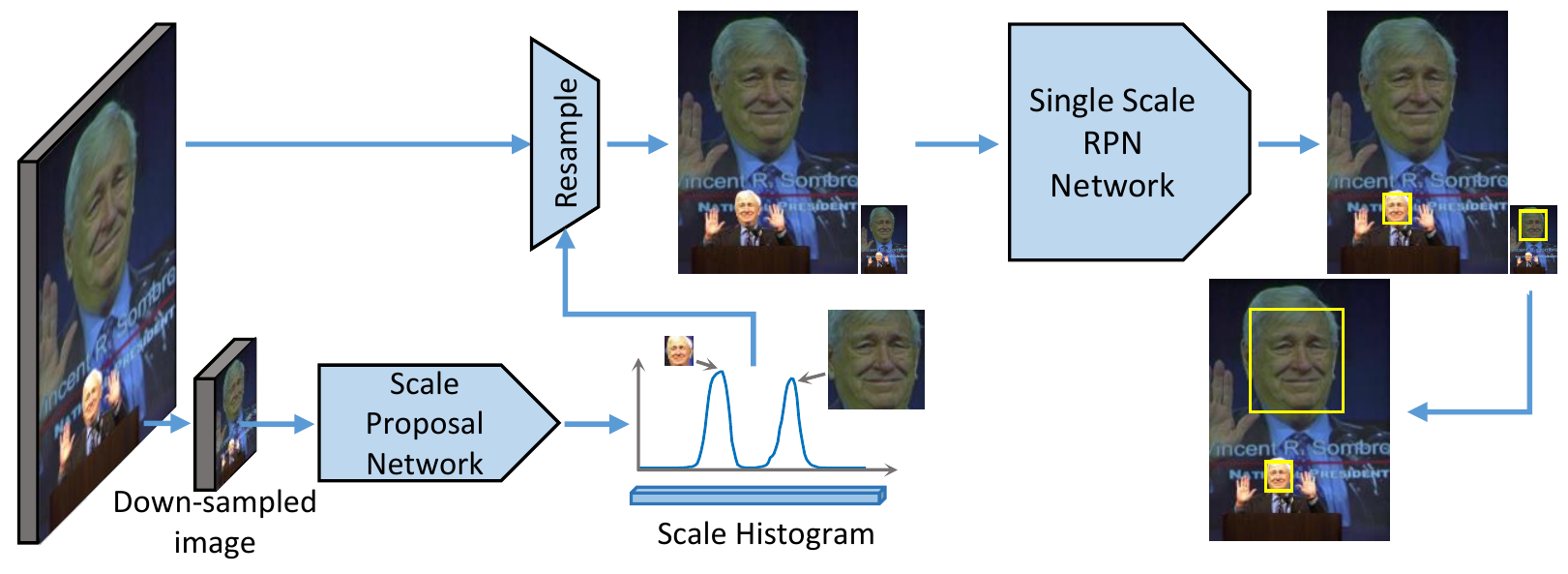}
\end{center}
   \caption{\textbf{The pipeline of Scale-Aware Face Detector.} Firstly, the input image is resampled to a small size and forwarded through Scale Proposal Network (SPN) to obtain Scale Histogram. The Scale Histogram encodes the possible sizes of faces in the image but it doesn't contain any location information. The SPN network needs little computation. Then the input image is resampled according to the Scale Histogram so that all the faces in the image fall in the coverable range of RPN. Computation can be reduced if the image contains only large faces. Finally, the resampled image set is individually detected for faces and the results are combined to obtain the final result.}
\label{fig:pipeline}
\end{figure*}

\section{Related works}
The CNN based face detection approaches emerged in 1990s ~\cite{vaillant1994original}. Some of the modules are still widely used, such as sliding window, multi-scale testing and the CNN based classifier to distinguish faces from background. \cite{rowley1998neural} shows that CNN achieves good performance for frontal face detection and \cite{rowley1998rotation} further extends it for rotation invariant face detection by training faces of different poses. Despite their good performance, they are too slow when considering the hardware of early years. 

One breakthrough in face detection is the Viola-Jones framework \cite{viola2004robust}, which combines Haar feature, Adaboost and cascade in face detection. It becomes very popular due to its advantages in both speed and accuracy. Many works have been proposed to improve the Viola-Jones framework and achieves further improvements, such as local features~\cite{zhang2007face,li2013learning,yang2014aggregate}, boosting algorithms~\cite{zhang2005multiple,li2004floatboost,huang2005vector}, cascade structure ~\cite{bourdev2005robust} and multi-pose \cite{li2002statistical,jones2003fast,huang2007high}.

The HOG based methods are firstly used in pedestrian or general object detection, such as the famous HOG \cite{dalal2005histograms} and deformable part model \cite{felzenszwalb2010object}. These methods achieves better performance than Viola-Jones based methods on standard benchmarks such as AFW~\cite{zhu2012face} and FDDB~\cite{jain2010fddb}, and progressively become more efficient, including \cite{zhu2012face,mathias2014face,yan2014fastest,ghiasi2015occlusion}.

The CNN based methods again become popular thank to their great performance advantages. Early works combine CNN based features with traditional features. \cite{ranjan2015deep} combines CNN with deformable part model and \cite{yang2015convolutional} combines CNN with channel feature \cite{dollar2014fast}. \cite{yang2015facial} predicts face part score map through fully convolutional networks and uses it to generate face proposals for further classification. \cite{li2015convolutional} proposes a CNN cascade for efficient face detection. This work is further improved in \cite{qin2016joint} with joint training. \cite{huang2015densebox} gives an end-to-end training version of detection network to directly predict bounding boxes and object confidences. \cite{farfade2015multi} shows that simple fine-tuning the CNN model from ImageNet classification task for face/background classification leads to good performance. In~\cite{chen2016supervised}, the supervised spatial transform layer is used to implement pose invariant face detection. Popular general object detection methods, such as Faster-RCNN \cite{ren2015faster}, R-FCN \cite{dai2016r}, YOLO \cite{redmon2015you} and SSD \cite{liu2015ssd} can also be used directly for face detection. Our proposed scale-aware face detection method is also a CNN-based method. However, it focuses on scale problem in face detection, in the way that, to our best knowledge, no one has ever explored yet. Our method is orthogonal to these CNN-based methods and they can benefit from each other.

There are some successful attempts on better handling of scale in object detection. They either construct stronger network structure by combining features from different depths of a network \cite{bell16ion} or directly predicting objects at different depth of a network \cite{cai16mscnn, liu2015ssd}. All of them share the same motivation. Intuitively, larger faces require a network with larger receptive field to be detected correctly, while smaller faces need a network with high resolution (and possibly smaller receptive field) to have it detected and localized correctly. But these methods have two major drawbacks. 
First, they fail to explicitly share feature between scales. These methods only share feature implicitly by sharing part of the convolution layers. The network still have to cover large scale variance, possibly needing more parameters to work well.
Second, in order to cover largest and smallest faces simultaneously in a single pass, the input image has to be large to prevent small faces from missing, even if the image doesn't contain small faces at all. This hurts speed considerably, and can be inferred from the FLOPs comparison in Figure~\ref{fig:motivation}. Both problems are tackled in SAFD.

\section{Scale-aware detection pipeline}

We propose SAFD that implicitly considers face scale variation. As illustrated in Figure~\ref{fig:pipeline}, our method consists of two stages, which disassembles face detection problem into two sub-problems: (1) global scale proposal and (2) single-scale detection. 
The goal of global scale proposal stage is to estimate the possible sizes of all the faces appearing in the image as well as assign a confidence score to each scale proposals. Then the image is scaled according to the scale proposals and detected for faces using single-scale RPN. If multiple scale proposals are generated in one image, it is scaled and detected for multiple times and results are combined to form the final detection result.

\subsection{Scale Proposal Network (SPN)}

We define scale proposals to be a set of estimated face sizes along with their confidences. The definition of face size is discussed in Section~\ref{section:gtprep}. In scale proposal stage, scale proposals are generated by Scale Proposal Network(SPN), a specially-designed convolutional neural network that aims at generating scale histogram with minimum human-introduced constraints. 

\begin{figure*}[t]
\begin{center}
   \includegraphics[width=0.6\linewidth]{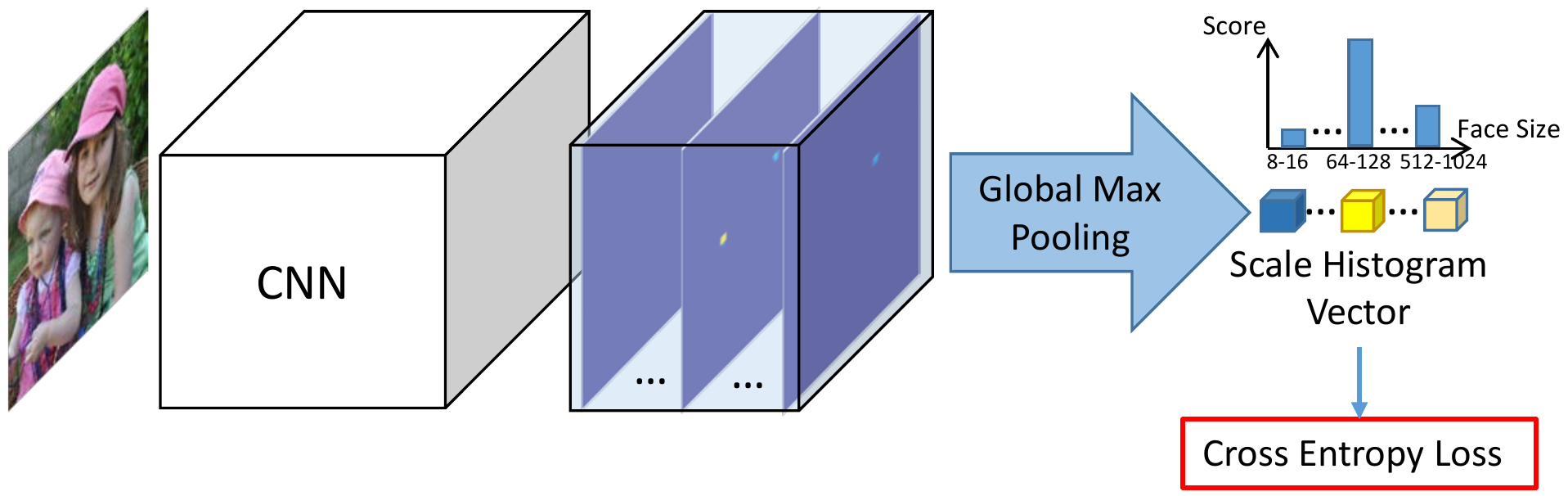}
\end{center}
   \caption{\textbf{The construction of Scale Proposal Network.} The SPN is a CNN with a global max pooling layer at its end so that it can produce a fixed-dimensional Scale Histogram Vector disregard the input size and face locations. Each element in Scale Histogram Vector represents the possibility of the presence of faces that have sizes within a certain range. During training, SPN only requires image-level supervision. }
\label{fig:scaleestimationnetwork}
\end{figure*}

The Scale Proposal Network is a fully convolutional network that has a global max-pooling layer after the last convolution layer for generating a fixed-length histogram vector from an input image of arbitrary size. Figure~\ref{fig:scaleestimationnetwork} shows the structure of Scale Proposal Network. It takes the down-sampled image as input, and produces a scale response heatmap (of size $w\times h\times n$). After global max-pooling the heatmap is reduced to a histogram vector of size $1\times 1\times n$, with each of its element corresponding to the probability of having faces of certain scale in the image. The histogram vector can be interpreted as a scale-vs-probability histogram. The output feature length is equal to the number of bins in the scale histogram. The histogram is normalized by Sigmoid function so that each element is within $[0,1]$ and represents probability.

The detailed explanation of scale histogram goes as follows. For a scale histogram with $n$ equally placed bins in log scale, with left edge corresponding to face size $s_0$ and right edge corresponding to face size $s_n$, the histogram vector $h$ is defined as:

\begin{equation}
h = [a_1, a_2, a_3, ..., a_n],
\end{equation}

\begin{equation}
\begin{split}
a_i = P(\exists {\rm x}|s^l_i \le {\rm log_2}({\rm size}({\rm x})) < s^r_i), \\
(i = 1, 2, ... ,n),
\end{split}
\end{equation}

\noindent where $d$ is the width of each bin in base-2 logarithmic scale, $d = (s_n - s_0)/n$,
$s^l_i$ and $s^r_i$ are the left and right edge of $i$th bin, so $s^l_i = s_0+(i-1)d$ and $s^r_i = s_0+id$. 
The ${\rm x}$ represents a face and ${\rm size}({\rm x})$ is the size of face ${\rm x}$.

In other words, $i$th histogram bin corresponds to faces whose sizes are within the following range:

\begin{equation}
[2^{s_0+(i-1)d}, 2^{s_0+id})
\end{equation}

With the network structure mentioned above, the global max-pooling layer essentially becomes a response aggregator, which discards location information and picks the maximum response of each histogram bin from all locations. This is a big advantage since it removes the location constraint that presents in standard RPN. The training process of RPN inherently holds the assumption that the response on the classification heatmap should be high if its projected position on input image is close to the center of an object. However, in SPN, the scale estimation response of a face can be at arbitrary location of the heatmap. Ignoring the location information helps the network to selectively learn highly representative features from faces and from context, even if the face is much larger or much smaller than the receptive field of the network. Moreover, this arrangement enables response from multiple face parts to contribute to scale estimation independently. Only the highest response will be selected, thus robustness can be improved. The training strategy for SPN is discussed in Section~\ref{section:globalsupervision}. 

\begin{figure}[t]
\begin{center}
  \includegraphics[width=0.8\linewidth]{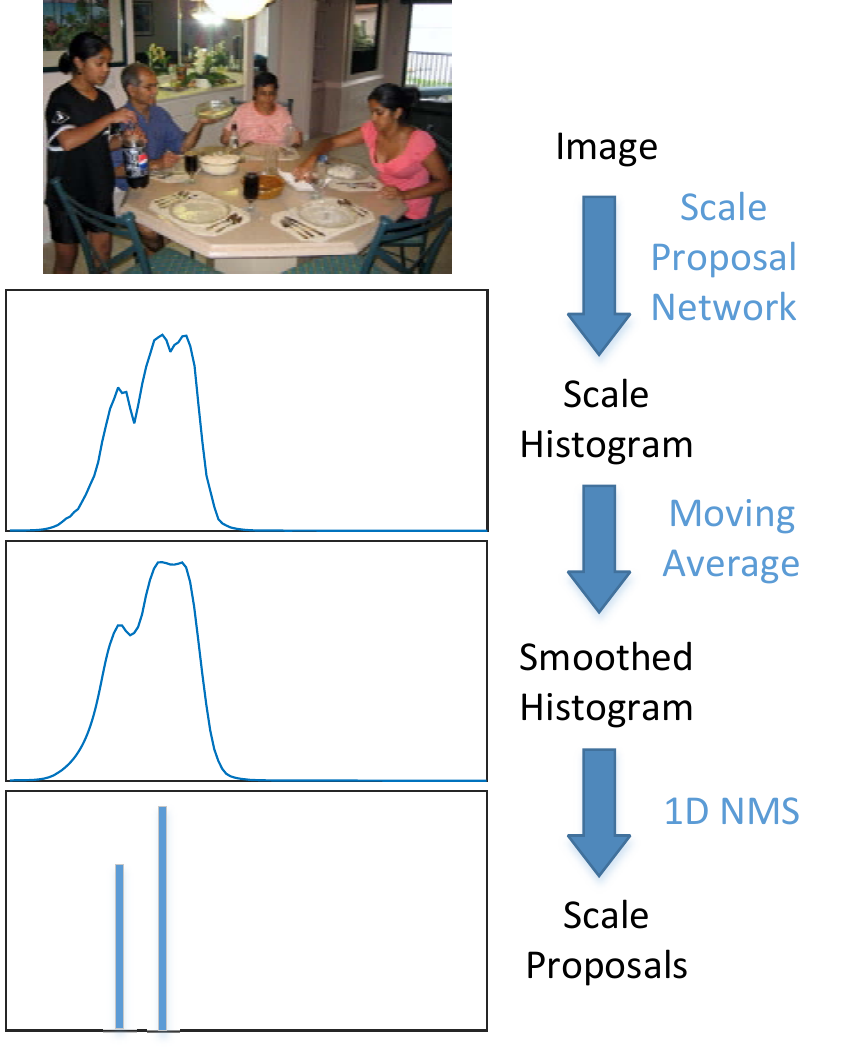}
\end{center}
   \caption{\textbf{Process of generating scale proposals from input image.} At first, the Scale Histogram of the image is generated by SPN. Then, the histogram is smoothed by moving average to reduce noise. Finally, non-maximum suppression is performed on smoothed histogram to obtain the final scale proposals. By using NMS, neighboring scale proposals can be efficiently combined to one proposal, which greatly saves computation. After NMS, only a few proposals left.}
\label{fig:histogram}
\end{figure}

\subsection{Scaling strategy generation}

There may be more than one face in an image. To save computation, we hope that faces that are close in size can be covered by detector in a single pass. Thanks to the high-resolution scale estimation generated by SPN, this can be implemented easily by non-maximum suppression (NMS).

When the estimated scale histogram has a large number of bins (e.g. 60 bins between face size of $2^3$ and $2^9$, with each bin having an interval of $2^{0.1}$), the histogram tends to be noisy. Moreover, the presence of a face in the image usually brings high response to its corresponding bin together with its adjacent bins, which makes it impossible to simply thresholding out the high-response proposals(Figure~\ref{fig:histogram}).

To extract useful signal from the histogram, the histogram is smoothed using moving average method with a window of half the length of the detector's covered range. This reduces high-frequency noise and spikes while retaining enough resolution. Then a one-dimensional NMS is applied to extract peaks from the smoothed histogram. The position of the peaks corresponds to face size while the heights of the peaks are regarded as their confidence scores. The window size for NMS is set to be slightly smaller than the cover range of the detector so it will not miss out useful signals (e.g. the scale response generated from another face). 

After NMS there are only a very small number of scale proposals left. Proposals that have a confidence higher than a threshold will be selected as final proposals and images are resized accordingly prior to detection. Although the above-mentioned strategy cannot guarantee to get the minimum number of scales per image, this sub-optimal solution can already achieve high recall rate while keeping number of final proposals small. 

\subsection{Single-scale RPN}

We adopt Region Proposal Network (RPN) as face detector in our pipeline, though any detector should behave similarly. The RPN is a fully convolutional network that has two output branches: classification branch and bounding box regression branch. Each branch may have one or many sub-branches, which handle objects of different scales. The reference box of each sub-branch is called \textit{anchor box}. The detailed information about RPN can be found in\cite{ren2015faster}.

Since the face size variation is already handled in the first stage, in this stage, we only use an RPN with one anchor. The largest detectable face size is set to be twice the size of the smallest detectable face. This configuration is enough to achieve high accuracy while keeping average zooms per image low and the RPN computationally cheap. The RPN we use is called \textit{Single-Scale RPN}, since it has only one anchor and has a narrow face size coverage.

\section{Implementation details}

\subsection{Global supervision} \label{section:globalsupervision}

The output histogram vector of SPN is directly supervised by sigmoid cross entropy loss:

\begin{equation}
L = -\frac{1}{N} \sum_{n=1}^{N}[p_n  \log \hat{p}_n + (1-p_n) \log (1-\hat{p}_n) ],
\end{equation}

\noindent where $N$ denotes the total number of bins, $\hat{p}$ is the histogram vector estimated by the network (normalized by sigmoid function), and $p$ is ground truth histogram vector.

Unlike the training process of RPN, no location information is provided to the SPN during training. What really happens during training is that, in each iteration, the gradient only back-propagates through the location with highest response. Although the SPN is trained from random initialization and the location selection may not always be correct especially in the first few iterations, it will be sticking to right location after thousand iterations' trial and error as long as the training data is sufficient. Owing to the fact that similar feature from irrelevant locations cannot be generalized to all the training samples, the SPN under global supervision will automatically learn features that can easily be generalized, as well as quickly rejecting features that are most likely to cause false scale proposals. 

\begin{figure}[t]
\begin{center}
   \includegraphics[width=0.8\linewidth]{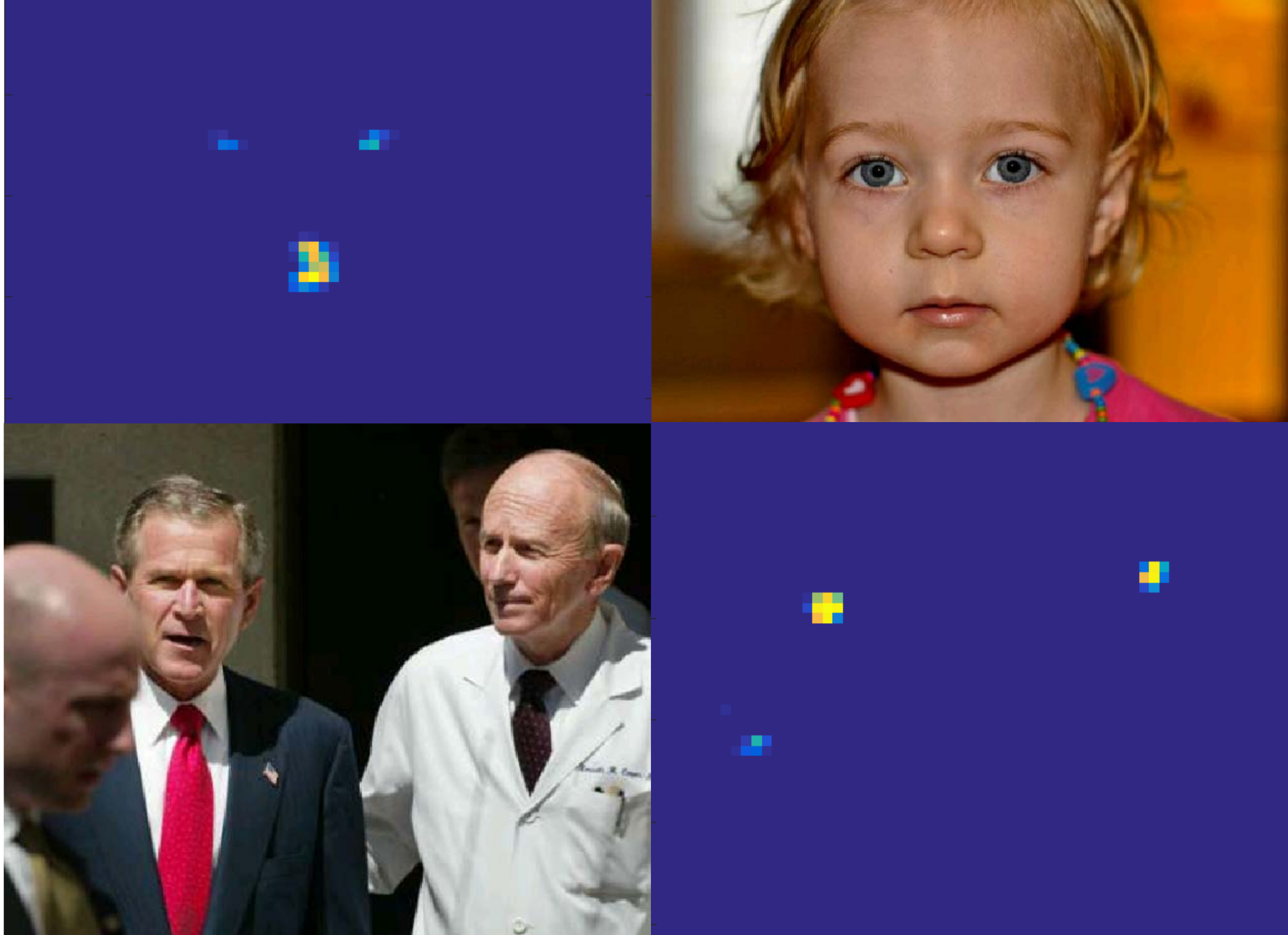}
\end{center}
   \caption{\textbf{Scale response map for face larger and smaller than the receptive field of SPN.} The upper-right face is significantly larger than the receptive field. Its corresponding response map on the upper left reveals facial landmark locations, which suggests that even if the face is larger that receptive field, SPN can still correctly recall it according to parts of faces. Also, although we don't supervise the locations of faces at SPN stage, the response map before global max pooling can still reveals some location information. }
\label{fig:receptivefield}
\end{figure}

No localization constraints is one of the desirable property of global supervision. When training fully-convolutional detectors or segmentation networks, the location of ground-truth samples are assigned on the heatmap using a set of strategies. These manually-assigned ground truths introduce strong constraints to the training process. One of the examples of those constraints is that, for RPN, the location on the heatmap must correspond to the same location on input image. By removing these constraints and allowing the network to learn to adjust to good features and suitable response formats itself, performance can be improved. One obvious benefit of global supervision is that this enables networks with small receptive fields to generate correct scale proposals for faces several times larger than the receptive field, thus reducing the need of deep networks. The SPN under global supervision can automatically generate scale proposal according to feature-rich facial parts, as shown in Figure~\ref{fig:receptivefield}. Another desirable property of global supervision is its inherent hard-negative mining nature. Global max-pooling always select highest response location for back propagation, thus highest response negative sample will always be selected in each iteration.

Although scale proposals can also be generated by a more complex, wide-range and single view detector such as a multi-anchor RPN, its speed cannot match SPN. 

\subsection{Ground truth preparation} \label{section:gtprep}

\noindent \textbf{Definition of bounding box.} The size of faces that used for generating ground truth histogram is defined to be the side length of the square bounding box. One problem regarding to this is that how to define the bounding box of a face and keep it consistent throughout the training samples. Noise in bounding box annotation can impair the performance of scale proposal network. Also, any misalignment of the bounding box between two stages can severely affect the performance. 

However, manual labeling of face bounding boxes is a very subjective task and prone to noise. So we prefer to derive bounding boxes from the more objectively-labeled 5-point facial landmark annotations using the transformation described below. Note that the bounding boxs we define are always square.

\begin{equation}
\begin{bmatrix} x_i^b \\ y_i^b \\s_i^b \end{bmatrix} = \begin{bmatrix} {\rm mean}(x_i^{l1},x_i^{l2},x_i^{l3},x_i^{l4},x_i^{l5}) +o_x\\ {\rm mean}(y_i^{l1},y_i^{l2},y_i^{l3},y_i^{l4},y_i^{l5}) +o_y \\{\rm std}(y_i^{l1},y_i^{l2},y_i^{l3},y_i^{l4},y_i^{l5}) * o_s \end{bmatrix}
\end{equation}

\noindent where $i$th landmark annotation $(x_i^{lk},y_i^{lk})$ corresponds to the location of \textit{left eye center}, \textit{right eye center}, \textit{nose}, \textit{left mouth corner} and \textit{right mouth corner} for $k = 1, 2,..., 5$ respectively. The corresponding bounding box is defined as $(x_i^b, y_i^b, s_i^b)$, where $(x_i^b, y_i^b$) is center location of the box and $s_i^b$ is its side length. $o_x$, $o_y$ and $o_s$ are offset parameters that are shared among all samples.

\noindent \textbf{Ground truth generation.} One of the most intuitive way to derive ground truth histogram from face sizes is by simply treating the histogram as multiple binary classifiers, setting the corresponding bin for each face to positive. But such nearest-neighbor approach is very prone to annotation noise even if the less-noisy annotation protocol is used. Though we managed to make nearest neighbor approach work on very large binning interval (e.g. $2^1$ bin width in log scale), its performance drops rapidly with the reducing of binning interval and can even prevent SPN from converging.

For the reasons above, we adopt a more stable approach for generating ground-truth histogram vector. For each ground truth face size $s$, we assign a Gaussian function: 

\begin{equation}
f(x) = e^{-\frac{(x-\log _2 s)^2}{2\sigma ^2}}.
\end{equation}

The target value for $i$th bin is sampled from $f(x)$:

\begin{equation}
a_i = f((s^l_i + s^r_i)/2).
\end{equation}

By doing so, the model is more immune to the noise introduced by imperfect ground truth since the Gaussian function provides a soft boundary. The selection of $\sigma$ mainly depends on the error distribution of ground truth and the window size of the detector. In our case, we use $\sigma = 0.4$ in all the experiments.

If more than one faces appear in a single image, the ground truth histogram is generated by doing element-wise maximum over the ground-truth histograms of each individual faces, which is coherent to the use of max-pooling layer.

\subsection{Receptive field problem}

Like all the fully-convolutional networks, in SPN the heatmap before global max-pooling has a limited receptive field. But unlike RPN, this receptive field limitation does not prevent the network from accurately estimating the size of faces that are many times larger than the receptive field. This is because some sub-regions from a large face contain enough information to inference the size of the whole face, as is described in Section~\ref{section:globalsupervision} and illustrated in Figure~\ref{fig:receptivefield}. Though the network we use has a receptive of $108 \times 108$ pixels, it can obtain sensible estimation of face sizes as large as $512 \times 512$ pixels.

\subsection{Training RPN}

The training of single scale RPN is straightforward. All the faces within the detectable range are regarded as positive samples and the faces outside the detectable range belong to negative samples.

\begin{figure*}[t]
\begin{center}
\subfigure[FDDB]{
   \includegraphics[width=0.32\linewidth]{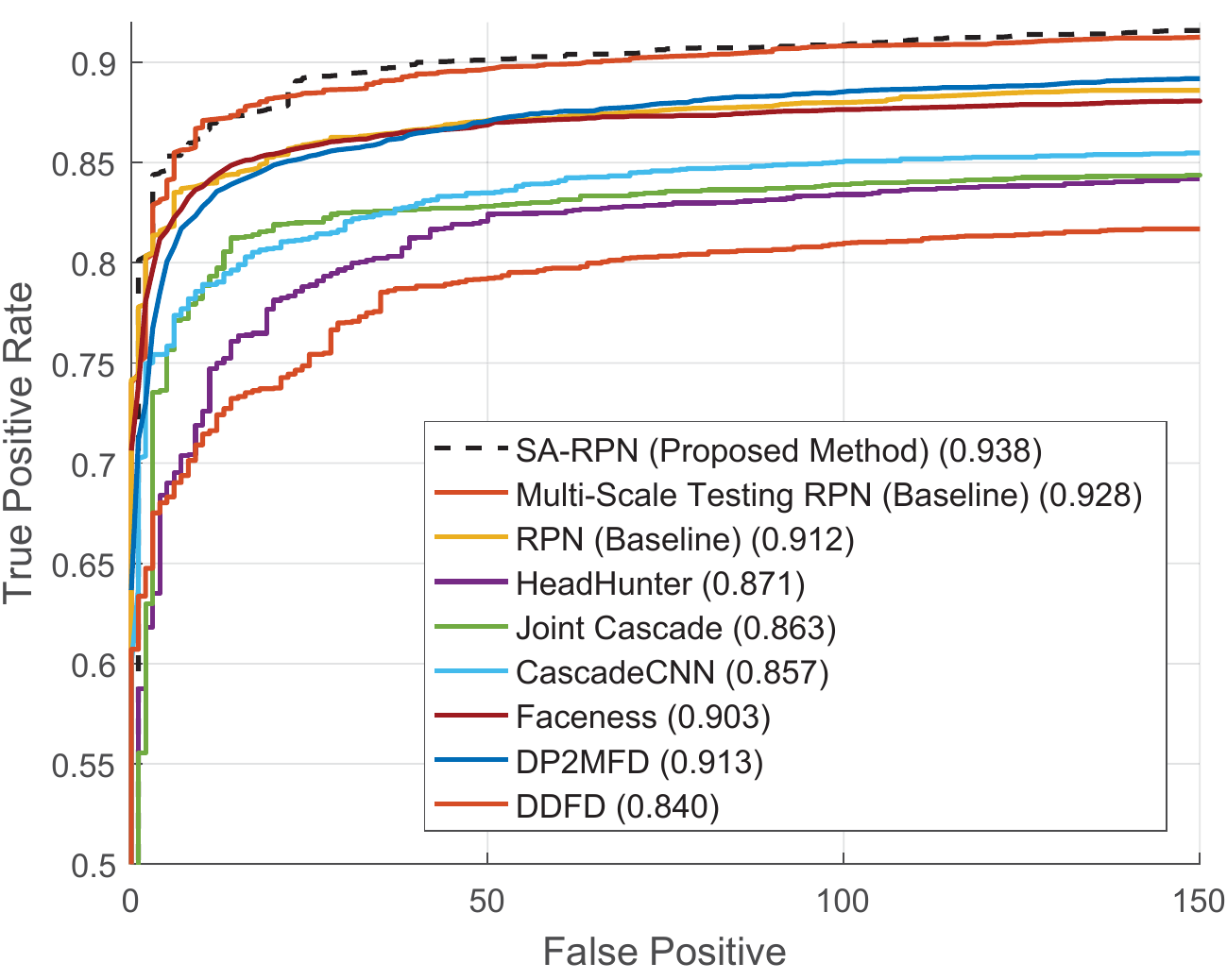}
 }
\subfigure[MALF - Whole]{
   \includegraphics[width=0.32\linewidth]{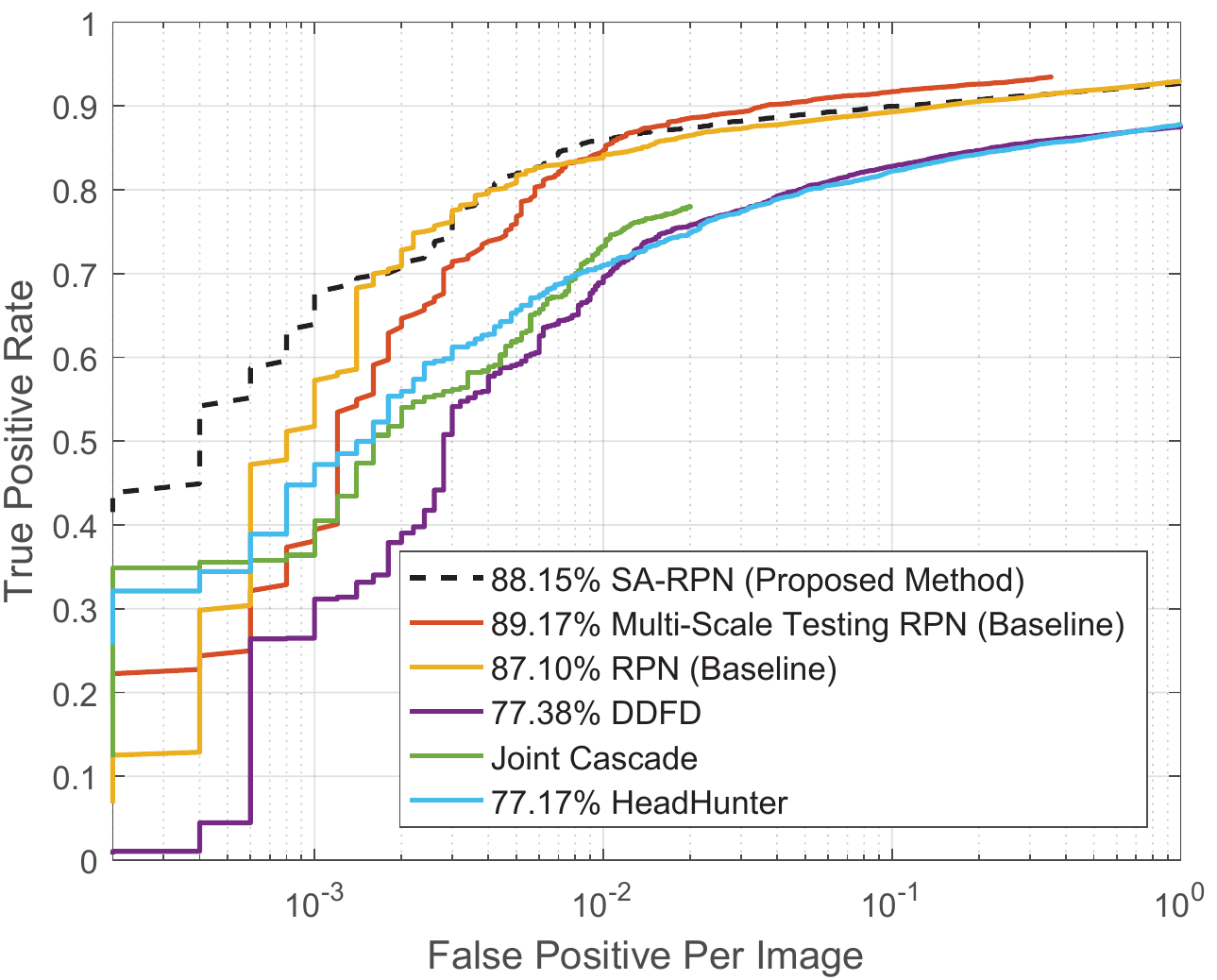}
 }
\subfigure[AFW]{
   \includegraphics[width=0.32\linewidth]{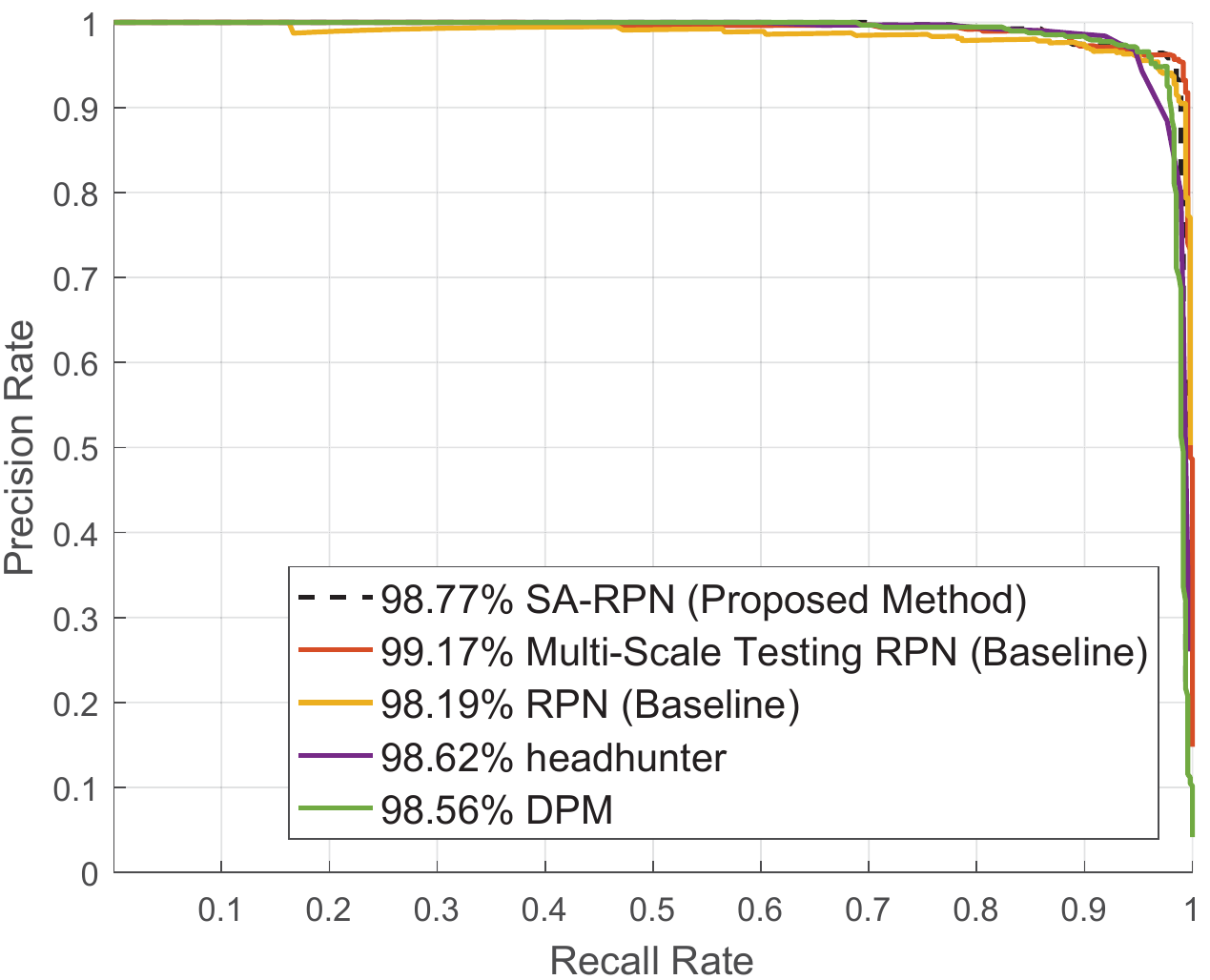}
 }
    
\end{center}
   \caption{\textbf{Comparison with previous methods on FDDB, MALF and AFW datasets.} The numeric metrics shown in figures are: (a) recall rate at 1000 false positives; (b) MALF proprietary ``mean recall rate''; (c) average precision. Best viewed in color.}
\label{fig:perfall}
\end{figure*}

\section{Experiments} \label{section:exp}
In this section, we will evaluate the performance of our pipeline on three face detection datasets: FDDB~\cite{fddbTech}, AFW~\cite{afw} and MALF~\cite{faceevaluation15}. We will also provide theoretical data for computational cost analysis. 

To make the experiment result comparable, we train both our model and other models under the same condition, using the same training data and the same network. The performance curves of our method along with several previous methods on each dataset are reported. Computational costs and time consumptions are listed and investigated. Extensive ablation experiments are conducted to validate the effectiveness of doing scale proposal prior to detection. 
In addition to overall performance, the performance of SPN is also separately evaluated.

\noindent \textbf{Training data overview. }For training samples, we collect about 700K images from the Internet, of which 350K contain faces. To improve the diversity of faces, we also include images from Annotated Facial Landmarks in the Wild (AFLW) dataset~\cite{kostinger2011annotated}. All the above-mentioned images are exclusive from FDDB, MALF and AFW datasets. 
For negative samples, we use both images from the Internet and COCO \cite{coco} dataset, excluding images with people 

All the easily-distinguished faces are labeled with 5 facial landmark points (left eye, right eye, nose, left mouth corner, right mouth corner) and bounding boxes were derived from the landmarks using the transformation described in Section~\ref{section:gtprep}. Faces and regions that were too hard to annotate were marked as ignoring regions. In the training of SPN, these regions will be filled with random colors before being fed into the network. In the training of RPNs, neither positive nor negative sample are drawn from these regions.

\noindent \textbf{Network structure. } Both SPN and RPN use a truncated version of GoogleNet, down to inception-3b. However, for SPN, the output channel of each convolution layer (within GoogleNet) is cut to 1/4 to further reduce computation. Table~\ref{table:gnetflops} shows the computational cost of each network. Batch normalization~\cite{ioffe2015batch} is used for both networks during training. 

\begin{table}
\begin{center}
\begin{tabular}{|c|c|c|}
\hline
\multirow{2}{*}{Layer} 	& \multicolumn{2}{| c |}{MFLOPs } 	\\
\cline{2-3}
 				& Full GoogleNet 	& 1/4 GoogleNet	\\
\hline\hline
conv1 			& 118  		& 30			\\
conv2 			& 360 			& 22			\\
inception(3a) 		& 171 			& 11			\\
inception(3b) 		& 203 			& 13			\\
feature128 			& 289 			& 72			\\
\hline
Total 				& 1141 		& 148 			\\
\hline
\end{tabular}
\end{center}
\caption{Architectures and computation analysis for Scale Proposal Network (1/4 GoogleNet) and Region Proposal network (full GoogleNet). All the data assume an input size of $224 \times 224 \times 3$. Batch Normalization layers are not shown and can be removed at test time. Auxiliary convolution layers are not shown for clarity.}
\label{table:gnetflops}
\end{table}

\noindent \textbf{Multi-scale testing RPN.} Each image is resampled to have long sides of $1414 \times 2^k (k = 0,-1,-2,-3,-4，-5)$. They are detected for faces respectively using the same RPN in our method. These intermediate results are combined to form the final result.

\noindent \textbf{Single view RPN.} A standard RPN that has 6 anchors to cover faces within the range of 8 to 512. The input image is always resized to have a long side length of 1414 pixels.

\begin{figure}
\begin{center}
   \includegraphics[width=0.8\linewidth]{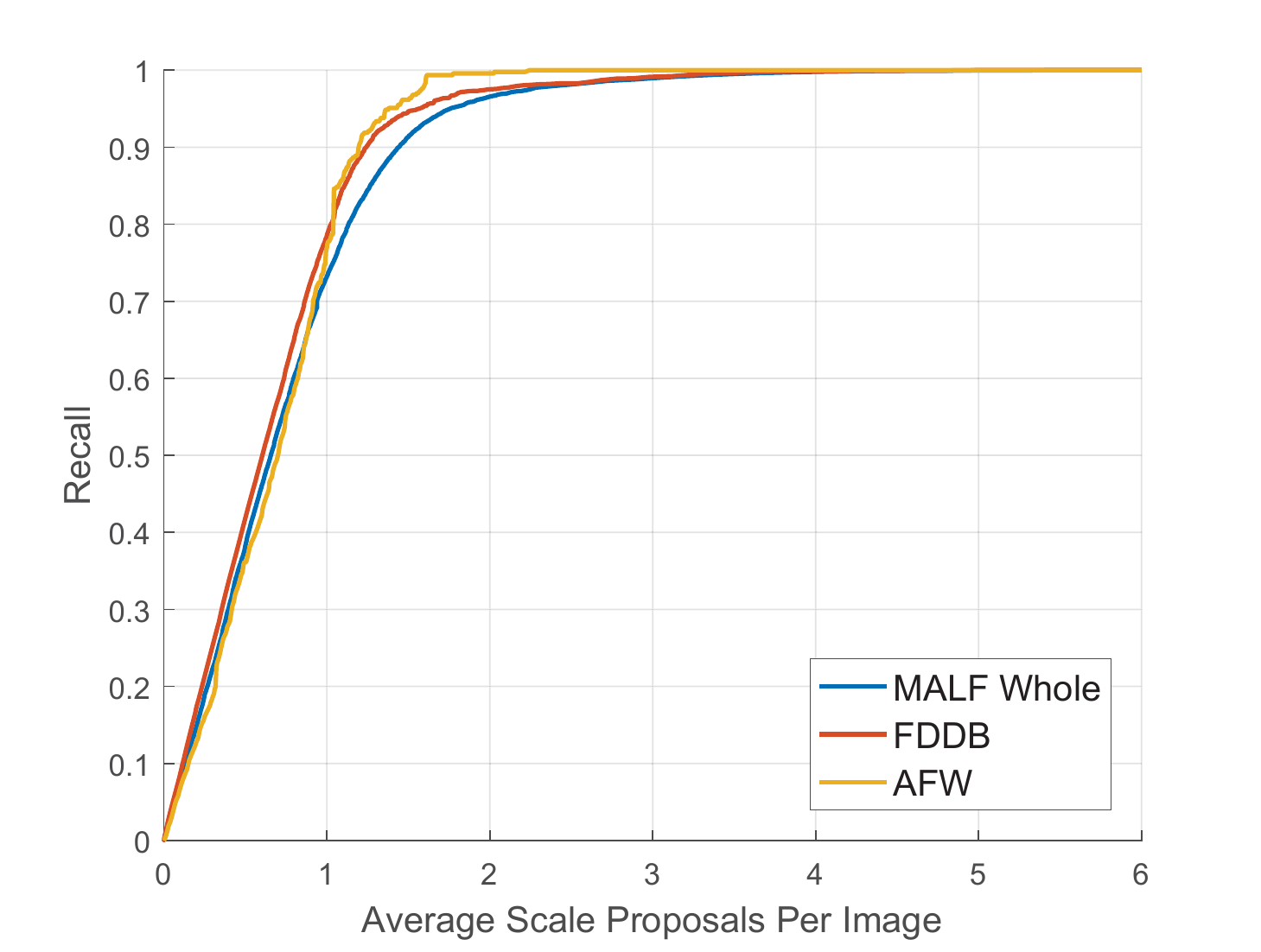}
\end{center}
   \caption{Recall-Average Scale Proposals Per Image curves of SPN on FDDB, MALF and AFW dataset.}
\label{fig:spnrecall}
\end{figure}

\begin{figure}
\begin{center}
   \includegraphics[width=0.8\linewidth]{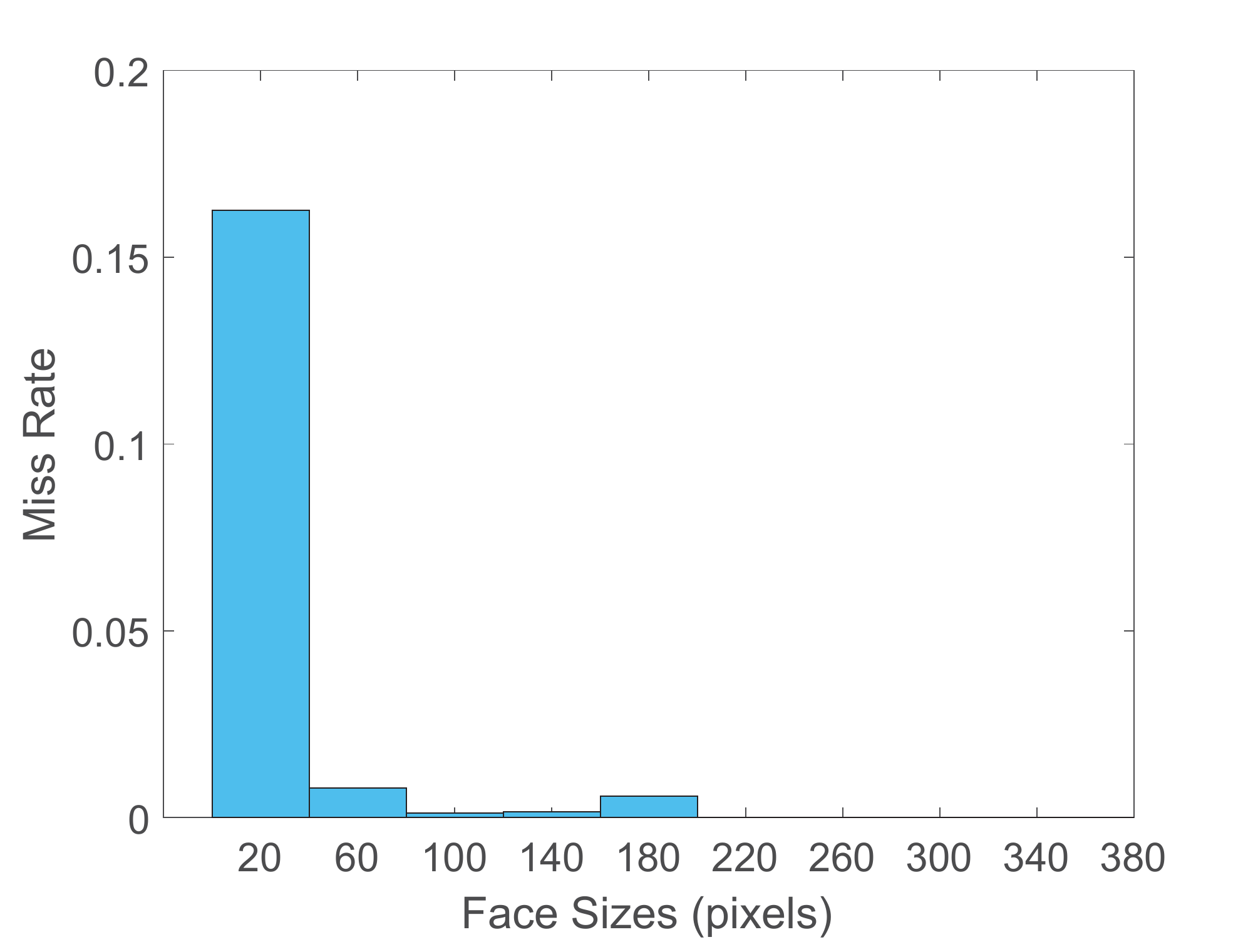}
\end{center}
   \caption{\textbf{The miss rate of SPN versus face size.} Miss rate is calculated as the proportion of faces not recalled in each bin. Evaluated on FDDB dataset.}
\label{fig:spnmrvsscale}
\end{figure}

\begin{table*}
\begin{center}
\begin{tabular}{|c|c|c|c|c|c|c|}
\hline
\multirow{2}{*}{Method} 	& \multicolumn{2}{| c |}{FDDB }	& \multicolumn{2}{| c |}{MALF Whole }	& \multicolumn{2}{| c |}{AFW } \\
\cline{2-7}
 				& MFLOPs 	& Time (ms) 		& MFLOPs 	& Time (ms) 		& MFLOPs 	& Time (ms) \\
\hline\hline
SA-RPN			&  441 + 2704	& 5.21 + 60.23 	& 437 + 8854 	& 11.87 + 190.15	& 432 + 6383	& 13.17 + 153.66	\\
MST-RPN			& 50240		& 754.75		& 49807			& 981.47 		& 49139		& 427.32	\\
RPN				& 33846		& 588.37		& 33554			& 549.68		& 33104		& 360.20	\\
\hline
\end{tabular}
\end{center}
\caption{Comparison of Scale-aware RPN (SA-RPN), multi-scale testing RPN (MST-RPN) and standard single-shot multi-anchor RPN (RPN) on computation requirements. The reported data are the average result for a single image.}
\label{table:flopsvsmodel}
\end{table*}

\subsection{Evaluation of scale proposal stage}

In this section, we first evaluate the performance of SPN separately from the whole pipeline. Since the scale proposal stage and detection stage essentially form a cascaded structure, any face that is missed by this stage will not be recalled by the detector. So, it is crucial to make sure that the scale proposal stage is not the performance bottleneck of the whole pipeline. We expect a high recall from this stage while keeping average resizes per image low. 

The SPN can handle faces within the range of $(2^3, 2^9)$, with a resolution of $2^{0.1}$. When testing, every image is resized so that its long side has a length of 448 pixels. A face is recalled only if its ground truth face size falls into the detectable range of detector (in our case 36-72 pixels) after being scaled according to the proposal.

We report the performance of SPN using Recall-Average Scale Proposals Per Image curves, as shown in Figure~\ref{fig:spnrecall}. We also analyze the SPN's performance on different face sizes. Figure~\ref{fig:spnmrvsscale} shows that most failures come from small faces while faces larger than receptive field can be handled well.

\subsection{Overall performance}

We benchmark our method on FDDB, MALF and AFW following the evaluation procedure provide by each dataset. For scale proposals, we discard proposals that have a confidence lower than a fixed threshold. Figure~\ref{fig:perfall} displays the performance of our method alongside with our baseline methods (Multi-Scale RPN, RPN) and state-of-the-art algorithms. Our method achieves best performance on FDDB and best accuracy in high confidence regions on MALF. The MALF dataset contains many challenging faces, having large face size diversity and a high proportion of small faces, which affect the recall rate of SPN and reduce the maximum possible recall of SAFD pipeline.

Though the chart does reveal that on MALF and AFW the SPN results in a drop on recall for low quality faces, our SAFD pipeline still outperforms previous methods. Moreover, the SA-RPN is several times faster than the slow but high recall multi-scale testing baseline and has fewer high-confidence false-positive detections. For multi-scale testing method, every image is detected in 6 different scales. Scale estimation reduces the average detection passes of each image, which can reduce the probability of getting false positives and improve speed.

The chart also shows that under the same condition, a single-shot multi-anchor RPN has significantly lower performance than SA-RPN and multi-scale testing RPN, which coincides with our expectation. Apart from the fact that such a RPN needs to fit to more diversified training data, the network has a receptive field of only 107 pixels, making it extremely hard to detect large faces correctly.

\subsection{Computational cost analysis}

In this section, we analyze the computational cost of SA-RPN along with baseline methods. Table~\ref{table:flopsvsmodel} shows the theoretical average FLOPs per image as well as empirical testing time on each database. Since the theoretical computation of CNN is proportional to the input image size (when taking padding area into account), the total FLOPs can easily be calculated by accumulating the input image size (in pixels) of CNNs on each forwarding pass. The test times contain system overheads so they are for reference purpose only. 

Unlike multi-scale testing RPN and standard RPN which has a fixed computational requirement on the same input size, the computational cost of our model is largely dependent on the content of images, which reveals in Table~\ref{table:flopsvsmodel} as large average FLOPs variance between datasets. But even on the worst-performing MALF dataset, our Scale Aware RPN can still outperforms baseline methods by a large margin in terms of speed. 

\section{Conclusion}

In this paper, we proposed SAFD, a two-stage face detection pipeline. It contains a scale proposal stage which automatically normalizes face sizes prior to detection. This enables computationally cheap single-scale face detector to handle large scale variation without using computationally expensive multi-scale pyramid testing. The SPN is designed to generate scale proposals. 
Our method achieves state-of-the-art performance on AFW, FDDB and MALF. The performance is similar to multi-scale testing based detectors but requires much less computation. 
The proposed method can also be applied to general object detection problems. Moreover, the SPN is essentially a weakly-supervised detector, which could be used to generate coarse region proposals and further improves speed. SPN can also share convolution layers with RPN to further reduce model size.

{\small
\bibliographystyle{ieee}
\bibliography{egbib}
}

\end{document}